\documentclass{article}

\usepackage[preprint]{neurips_2021}

\usepackage[utf8]{inputenc} 
\usepackage[T1]{fontenc}    
\usepackage{hyperref}       
\usepackage{url}            
\usepackage{booktabs}       
\usepackage{amsfonts}       
\usepackage{nicefrac}       
\usepackage{microtype}      
\usepackage{xcolor}         
\usepackage{subcaption}
\usepackage{graphicx,caption}
\usepackage{pythonhighlight}
\usepackage[rawfloats=true]{floatrow}
\usepackage[nocompress]{cite}

\title{DeepSim: A Reinforcement Learning Environment Build Toolkit for ROS and Gazebo  }

\author{%
  Woong Gyu La\qquad Lingjie Kong\qquad Sunil Muralidhara\qquad Pratik Nichat \\
  \\
  Amazon Web Services \\
  \\
  \texttt{\{woong,lingjik,murasuni,pnichat\} @ amazon.com} \\
}

\begin{document}

\maketitle

\begin{abstract}

We propose DeepSim \footnote{\url{https://github.com/aws-deepracer/deepsim}}, a reinforcement learning environment build toolkit for ROS and Gazebo.
It allows machine learning or reinforcement learning researchers to access the robotics domain and create complex and challenging custom tasks in ROS and Gazebo simulation environments.
This toolkit provides building blocks of advanced features such as collision detection, behaviour control, domain randomization, spawner, and many more.
DeepSim is designed to reduce the boundary between robotics and machine learning communities by providing Python interface.
In this paper, we discuss the components and design decisions of DeepSim Toolkit.

\end{abstract}

\section{Introduction}

Recently, reinforcement learning (RL) research has made multiple breakthroughs using simulation platforms such as Arcade Learning Environment \cite{ale2013}, VizDoom \cite{wydmuch2018vizdoom}, MuJoCo \cite{6386109}, Malmo \cite{10.5555/3061053.3061259}, and many others.
However, these platforms started to fall behind in meeting the current demand as benchmarks on these platforms are becoming less differentiable.
Researchers are looking for more practical and challenging environments to push the boundary of RL research.

In robotics communities, ROS \cite{quigley2009ros} and Gazebo \cite{koenig2004design} are used as foundation layers to build robotic control frameworks in both simulation and real world applications. 
For a simple robot simulation, the current Gazebo and ROS stack satisfy the researchers' needs.
However, since the Gazebo functionalities exposed to ROS are very primitive, building more complex tasks in Gazebo requires a development of C++ extensions.
Considering machine learning (ML) communities are more familiar with Python, the requirement of the C++ extension development becomes a barrier to them.
With the support of advanced features similar to Gazebo C++ plugin extension through a Python interface, it allows ML or RL researchers more approachable to Gazebo and ROS foundation work.

To bridge robotics and ML communities, we present DeepSim, a reinforcement learning environment build toolkit for ROS and Gazebo. 
It provides the building blocks of advanced features such as collision detection, behaviour controls, domain randomization, spawner, and many more to build a complex and challenging reinforcement learning environment in ROS and Gazebo simulation environments with Python language.

\section{Related Work}

In this section, we describe the related work of DeepSim. 
Robotics communities grew with Gazebo and ROS platform. 
DeepSim framework is built to bring ML communities to the robotics domain to solve the complex and practical problems with ML. 
It allows the ML researchers to create new tasks in Gazebo and ROS platforms with minimum efforts.

\subsection{Gazebo}

Gazebo \cite{koenig2004design} is an open-source 3D simulator for robotics applications. Gazebo was initially a component of the Player Project from 2004 to 2011.
In 2012, Open Source Robotics Foundation (OSRF) started supporting Gazebo as an independent project.
 
Gazebo integrates physics engines such as Open Dynamics Engine (ODE) \cite{smith2005open}, Bullet \cite{coumans2021}, Simbody \cite{sherman2011simbody}, and Dynamic Animation and Robotics Toolkit (DART) \cite{lee2018dart}.
A physical model described by Simulation Description Format (SDF) or Unified Robotic Description Format (URDF) file using XML format can be loaded by each of these physics engines.
Gazebo also utilizes OGRE as its rendering engine to render robot and environment models, and to process the visual sensors.

Gazebo provides various types of sensors such as camera, lidar, and many others that simulate the existing physical sensors.
Instantiated sensors listen to world state changes from a physics simulator and output the results.
Also, Gazebo allows users to come up with their own world, model, sensor, system, visual, and GUI plugins by implementing C++ Gazebo extensions.
This capability enables users to extend the simulator further into more complex scenarios.

\subsection{ROS}

Robot Operating System (ROS) \cite{quigley2009ros} is an open-source software framework for robot software development maintained by OSRF. 
ROS is a widely used middleware by the robotics researchers to leverage the communication between different modules in a robot and between different robots, and to maximize the re-usability of robotics code from simulation to the physical devices.  

ROS allows to run different device's modules as a node and provide multiple different types of communication layers between the nodes such as service, publisher-subscriber, and action communication models to satisfy different purposes. 
This allows robotics developers to encapsulate, package, re-use each of the modules independently. 
Furthermore, it allows each module to be used in both simulation and physical device without any modification.

\section{DeepSim Toolkit}

\begin{figure}
  \centering
  \includegraphics[width=\textwidth]{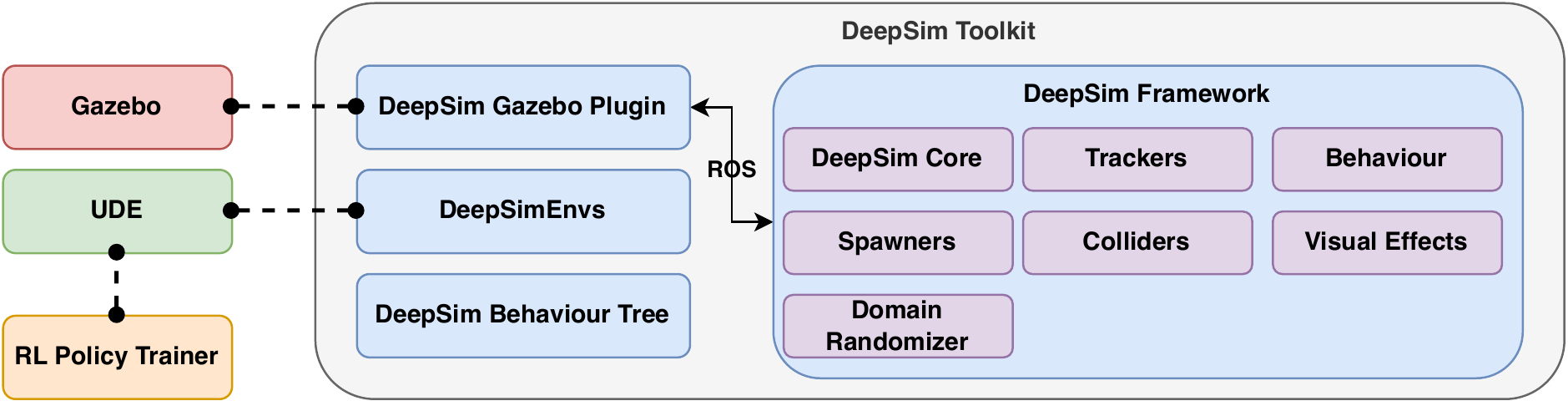}
  \caption{DeepSim Toolkit's high-level component architecture demonstrating the integration with external tools such as \emph{Gazebo} and \emph{Unified Distributed Environment} (UDE).}
  \label{fig:deepsim}
\end{figure}

In this section, we describe the design of our proposed toolkit. 
DeepSim (Figure \ref{fig:deepsim})  consists of 4 different components, DeepSimEnvs, Gazebo Plugin, DeepSim Framework, and Behaviour Tree to assist in building complex tasks in the ROS and Gazebo platform with Python. 
Each component is explained in more detail in the following sections.

\begin{figure}
  \centering
  \includegraphics[width=0.8\textwidth]{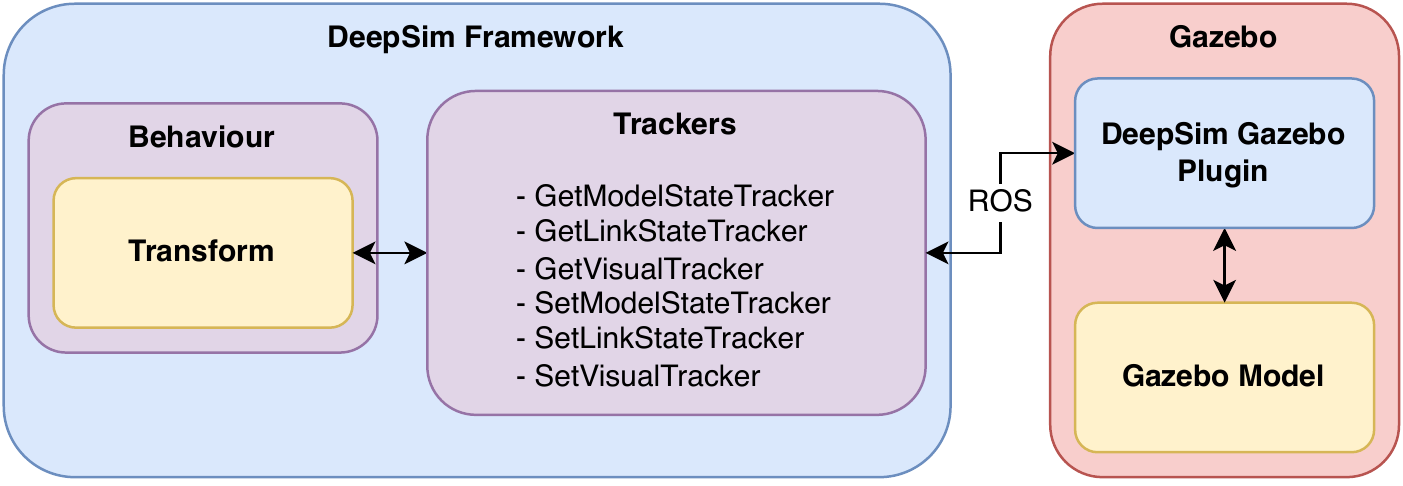}
  \caption{Illustration of data flow from Gazebo model to \emph{behaviour}'s \emph{transform}.}
  \label{fig:behaviour}
\end{figure}

\subsection{DeepSim Gazebo Plugin}
\label{section:deepsim_gazebo_plugin}

OSRF provides a bridge between Gazebo and ROS with \emph{gazebo\_ros} plugin package. 
However, \emph{gazebo\_ros} has a critical limitation that it can only process single retrieval or modification request of one model or link state at a time due to the communication using \emph{service} model \cite{quigley2009ros} over TCP.
Therefore, in order to \emph{get} or \emph{set} multiple model or link states, the time linearly increases with the number of \emph{get} and \emph{set} requests due to the network overhead and the nature of synchronous process in \emph{gazebo\_ros} plugin.
DeepSim Gazebo Plugin overcomes this limitation by introducing multi-state \emph{set} or \emph{get} services through a single request (see Appendix \ref{appendix_gazebo_get_set} for the usage comparison). 
DeepSim Gazebo Plugin is designed to process an arbitrary number of model states or link states in a single request. 
This design minimizes the network overhead regardless of the number of state change requests.

As illustrated in Figure \ref{fig:behaviour}, DeepSim Gazebo Plugin component provides the data communication between DeepSim Framework and Gazebo in primitive form by using ROS communication layer. 
The component provides the information from Gazebo through both \emph{service} and \emph{topic} model. 
It publishes all model, link, and visual states' information periodically through publisher-subscriber \emph{topic}s \cite{quigley2009ros}. 
Lastly, all model, link, and visual states' modification can be requested through \emph{service} request. 

\subsection{DeepSim Framework}

DeepSim Gazebo Plugin supports retrieval and modification of simulation raw state information such as model, link, and visual state data.
However, in many cases, solely simulation state information retrieval and modification are not sufficient to create a challenging task with a complex simulation environment setting.
In order to create more complex tasks, it often requires functionalities including but not limited to fine tuned object movement control, dynamic spawning of objects, waypoint navigation, and segmentation map.
From simple primitive state data retrieved with DeepSim Gazebo Plugin, DeepSim Framework provides advanced high-level features such as behaviour control, collision detection, spawners, domain randomization and many more to support these needs.

\subsubsection{DeepSim Core}

DeepSim Core contains basic game mathematics modules and primitive Gazebo entity modules to provide a foundation for advanced game engine features.
It can also be used for any custom usage by the users. 
DeepSim Core provides frequently used mathematics such as \emph{vector}, \emph{quaternion}, and \emph{euler}. 
For object management and Gazebo synchronization, \emph{link\_state}, \emph{model\_state}, \emph{point}, \emph{pose}, \emph{twist}, \emph{color}, \emph{material}, and \emph{visual} modules are provided. 
Some of the advanced game mathematics such as \emph{ray}, \emph{plane}, and \emph{frustum} are provided to support advanced game features like ray-casting, culling, or creating segmentation map. 
Lastly, DeepSim Framework provides \emph{lerp} and \emph{lerp\_angle} for quick linear interpolation functionalities, which are often used in camera or model movement control.

\begin{figure}
\floatsetup{footposition=caption, heightadjust=object}
\begin{floatrow}
\ffigbox{%
\begin{minipage}{0.50\textwidth}
  \centering
  \captionsetup{width=1.0\linewidth}
  \includegraphics[width=1.0\linewidth]{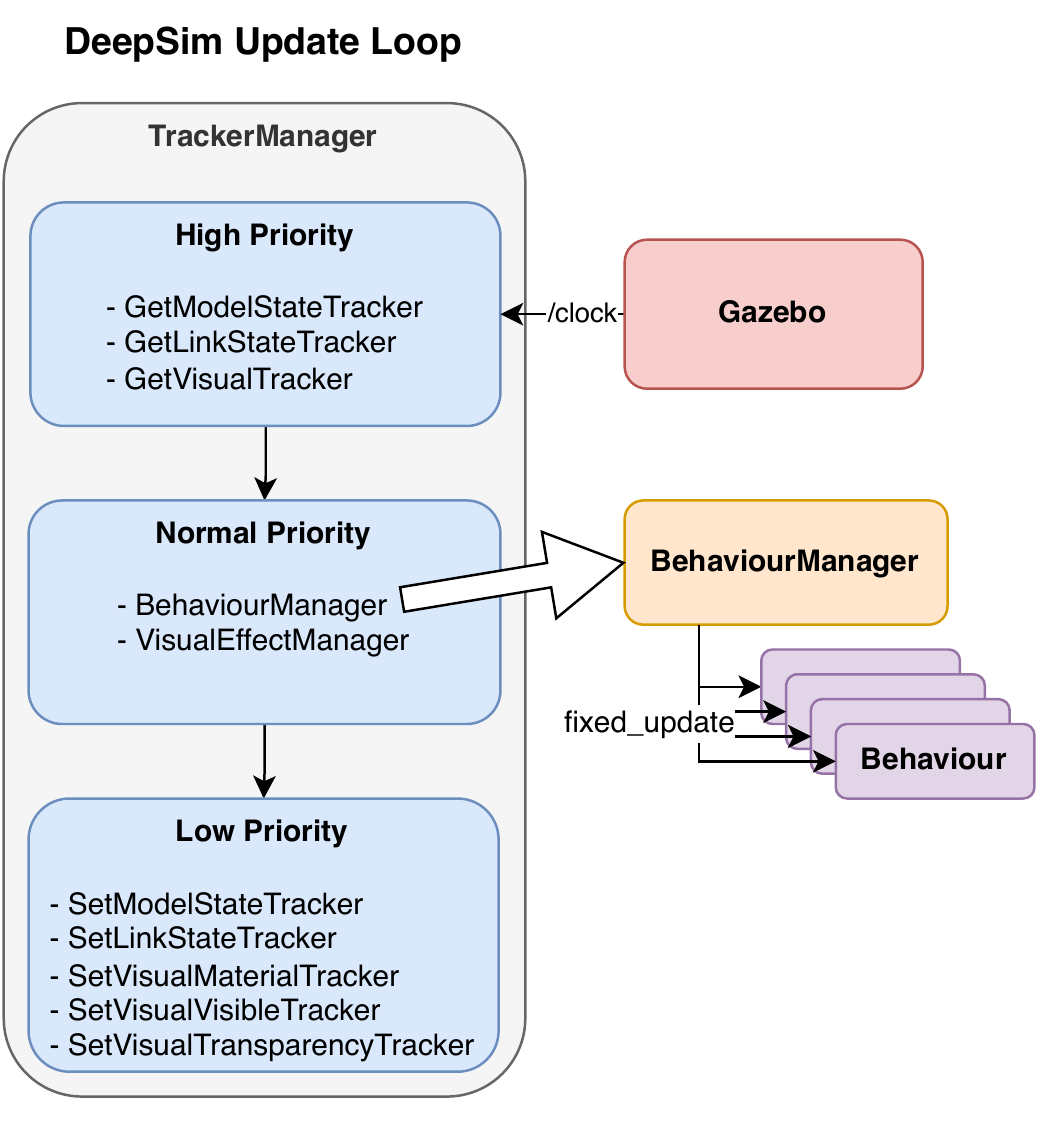}
  \caption{DeepSim Framework's update loop starting from Gazebo's simulation clock tick update.}
  \label{fig:loop}
\end{minipage}%
}{%
}
\ffigbox{%

\begin{minipage}{0.45\textwidth}
  \centering
  \captionsetup{width=1.0\linewidth}
  \includegraphics[width=1.0\linewidth]{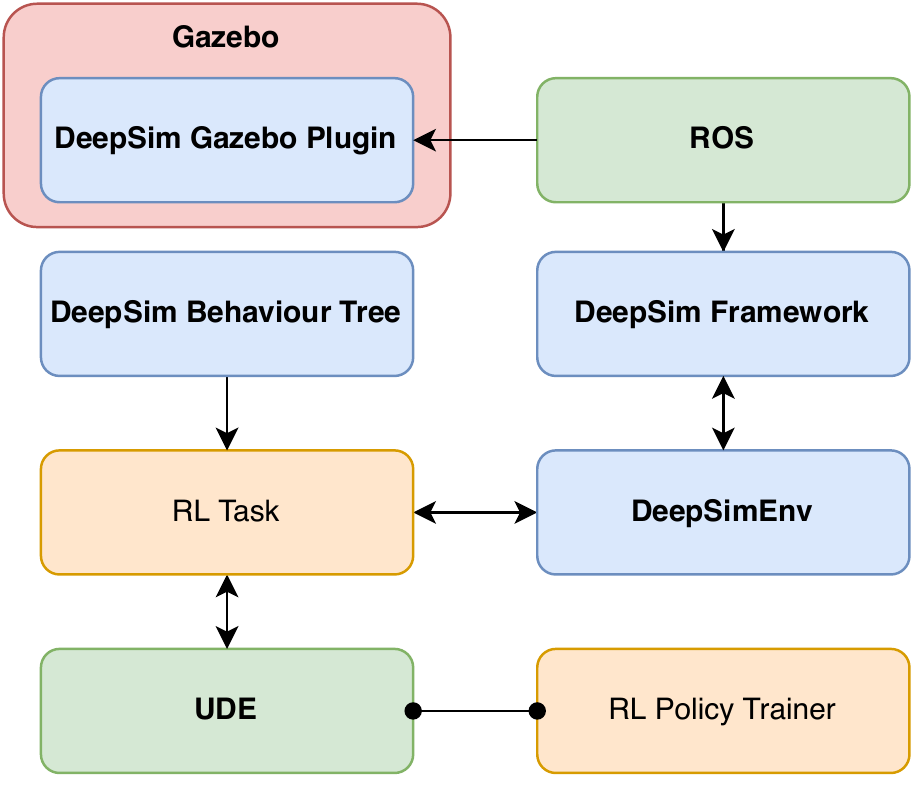}
  \caption{Complete DeepSim component stack illustrated from Gazebo simulator to agent policy trainer.}
  \label{fig:stack}
\end{minipage}%
}{%
}
\end{floatrow}
\end{figure}

\subsubsection{Trackers}

The tracker is an object that is registered under the tracker manager.
Each tracker gets invoked per Gazebo simulation time-step update.
Tracker manager subscribes to \emph{clock} topic to get a callback invoked on every simulation time-step update.
During the callback, the tracker manager invokes \emph{update\_tracker} method to every tracker that it manages.
As illustrated in Figure \ref{fig:loop}, the callback is called based on the level of priority group, HIGH, NORMAL, and LOW. The trackers in the HIGH priority group are called initially, then the trackers with the NORMAL priority group, and the trackers with the LOW priority group lastly.

The tracker manager is registered with default getter and setter trackers for link, model, and visual states in Gazebo.
Getter trackers retrieve all information from Gazebo at the time of \emph{update\_tracker}.
Setter trackers collect all \emph{set} requests for models, links, and visuals from previous update till the next \emph{update\_tracker} call, and make a single request to Gazebo simulator through DeepSim Gazebo Plugin.
Getter trackers, which synchronize all the latest information in Gazebo to DeepSim Framework, are part of the HIGH priority group.
Setter trackers, which will synchronize back all the information modified back to Gazebo at the end of the update loop, have LOW priority.
Consequently, the update loop starts by synchronizing all the information from Gazebo simulator, and all the updates made during the current update cycle will be synchronized back to Gazebo at the end of the update loop.

\subsubsection{Behaviour}

Behaviour represents a basic object entity (or \emph{model} in Gazebo term) in DeepSim Toolkit.
For each of \emph{behaviour} objects, a \emph{transform} is instantiated to maintain and synchronize the properties with Gazebo simulator through getter and setter trackers.
All states, updated in Gazebo simulator, are automatically synchronized to \emph{transform}, and any new properties configured with \emph{transform} are synchronized back to Gazebo simulator as shown in Figure \ref{fig:behaviour}.
The \emph{behaviour} also takes in a \emph{spawner} to allow the user to control the life-cycle of the object through \emph{behaviour} module.
All \emph{behaviour}s are automatically managed by \emph{behaviour\_manager}. The \emph{behaviour\_manager} is implemented in singleton pattern to provide the access to \emph{behaviour} objects in any place in the code.

The \emph{behaviour} must provide the name and the tag.
The name and the tag can be used to retrieve the \emph{behaviour} object from \emph{behaviour\_manager}.
This is especially useful when the user tries to operate on multiple entities with the same tag.
Lastly, \emph{behaviour} supports two types of update, \emph{update} and \emph{fixed\_update}.
The \emph{update} is either invoked manually or through the environment step, but \emph{fixed\_update} is invoked every Gazebo simulation time-step as shown in Figure \ref{fig:loop}.
Therefore, the behaviour controls can be updated with \emph{update} method, and the update that requires smooth interpolation such as model or camera movement, can be updated using \emph{fixed\_update}.

\subsubsection{Colliders}

The collisions are automatically processed by Gazebo simulation through an integrated physics engine, but such information is not accessible unless the user creates a custom Gazebo plugin to retrieve such information. 
Compared to the actual collision simulation, the access to manual collision detection functionality is often useful to create complex tasks in the simulation environment such as checking model collision with way-points to track the progress, offtrack detection, or creating segmentation map. 
DeepSim Framework provides frequently used 2D and 3D collider types for general use. 
For 2D, rectangle, circle, and geometry colliders are supported. 
Moreover, box and sphere colliders are supported in 3D.

Colliders can be attached to a \emph{transform} to track and automatically update its pose. 
Colliders also allow to set pose offset to provide more complex pose configuration. 
Colliders provide three main test methods, \emph{intersects}, \emph{contains}, and \emph{raycast}. 
The \emph{intersects} returns the flag of whether the collider intersects with a given collider or point.
The \emph{contains} returns the flag of whether the collider contains a given collider or point. 
Lastly, \emph{raycast} provides \emph{hit} information with a given \emph{ray} object. 
2D colliders are computed by using Shapely \cite{shapely}, and 3D colliders are computed by employing the mathematics modules from DeepSim Core.

\subsubsection{Spawners}

DeepSim Framework requires a \emph{spawner} implementation for each of \emph{behaviour} types to control the life-cycle of relevant models in Gazebo simulator.
The spawner must implement \emph{spawn} and \emph{delete} functions to create and delete the model in the Gazebo simulator respectively.
DeepSim Framework also provides two helper modules for the \emph{spawner} --- \emph{GazeboXMLLoader} and \emph{GazeboModelSpanwer}.
\emph{GazeboXMLLoader} parses and loads SDF or URDF files in XML or Xacro format.
Further \emph{GazeboModelSpanwer} provides a simple interface to spawn and delete SDF or URDF models in Gazebo simulator.

\subsubsection{Visual Effects and Domain Randomizers}

DeepSim Framework provides a feature to manage visual effects.
All visual effect implementations are automatically managed by the effect manager.
When an effect is attached, it is automatically added to the effect manager, and the effect manager is responsible to update each of the effect implementations until the effects are detached.
Effect manager is implemented as a tracker for its callback function to be invoked every Gazebo simulation time-step update.
Per update, the effect manager will propagate the \emph{update} call to every effect implementation that it manages.
DeepSim Framework provides two out of the box visual effects --- blink and invisible.
The user can further extend the abstract effect class to create their own custom visual effects.

Domain randomizer feature is supported by DeepSim Framework to apply domain randomization to the model that is spawned in Gazebo simulator.
DeepSim Framework supports two types of domain randomizer: model visual randomizer and light randomizer. Model visual randomizer allows to randomize the color in different levels such as model, link, and visual.
Also, it supports the color range to choose from, the number of units to randomize for each randomization step, and the filters to apply randomization.
The light randomizer also supports the range of color to choose with attenuation ranges, constant, linear, and quadratic values.
DeepSim Framework allows the user to extend and create their own custom randomizer by subclassing an abstract randomizer interface.

\subsection{DeepSim Behaviour Tree}

\begin{figure}
  \centering
  \includegraphics[width=0.97\textwidth]{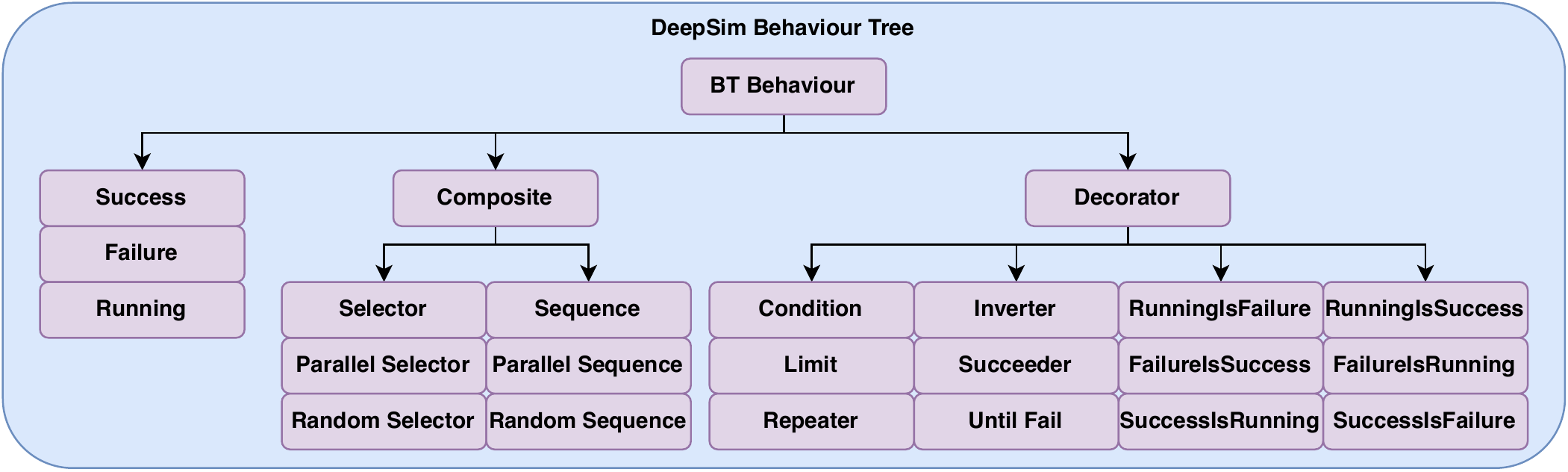}
  \caption{The behaviour tree nodes provided by DeepSim Behaviour Tree component.}
  \label{fig:deepsim_btree}
\end{figure}

Behaviour tree is popularly used in the game industry to model the behaviour of an autonomous agent or virtual entity in simulations or games \cite{btree2018, colledanchise2016behavior}.
Behaviour tree overcomes many limitations from Finite State Machine (FSM) approach in terms of maintainability, scalability, and re-usability.
While behaviour tree is popularly used in the game industry, not many behaviour tree implementations are available in Python language.
Based on our knowledge, there are no such behaviour tree implementations in Python language that are light-weighted and extendable.

DeepSim Behaviour Tree is provided as a component of DeepSim Toolkit to support advanced behaviour designing for the agents or object entities.
The component is very light-weighted and the nodes can be extended to create new reusable behaviour nodes.
As DeepSim Behaviour Tree is a stand-alone component, which does not depend on any other DeepSim Toolkit's components, DeepSim Behaviour Tree can be solely installed in any Python environment without other DeepSim Toolkit's components.

Comparable to other behaviour tree implementations, DeepSim Behaviour Tree also traverses and executes the nodes from the root with \emph{tick} method call.  
For every \emph{tick} operation, it returns one of four status --- SUCCESS, FAILURE, RUNNING, and INVALID, which indicates the execution was successful, failed, running, and invalid respectively.
\emph{BT Behaviour} represents a base node of the behaviour tree, and all behaviour tree nodes must be derived from \emph{BT Behaviour}.
As illustrated in Figure \ref{fig:deepsim_btree}, DeepSim Behaviour Tree also provides some useful extensions for each of the behaviour tree node types --- Leaf, Decorator, and Composite.
For further custom behaviour outside of what is provided, users can extend DeepSim Behaviour Tree nodes according to their needs.

\subsubsection{Leaf and Decorator}
Leaf node is a type of behaviour tree nodes without any child.
It executes and returns its status back to its parent node.
The leaf node often represents an entity's action or its status check.
DeepSim Behaviour Tree provides three leaf nodes out of the box --- \emph{Success}, \emph{Failure}, and \emph{Running}.
As their name states, on execution, the nodes just return the status that match with their names, where \emph{Success} always returns SUCCESS, \emph{Failure} always returns FAILURE, and lastly \emph{Running} always returns RUNNING.

Decorator node is a node that only has a single child node.
It is often used to manipulate the return status of its child node, or control the behaviour of its child node.
DeepSim Behaviour Tree provides a handful of pre-defined decorator nodes such as \emph{Condition}, \emph{Limit}, \emph{Repeater}, \emph{Inverter}, \emph{Succeeder}, \emph{UntilFail}, and status manipulators.
\emph{Condition} returns SUCCESS if the child node returns the targeted status otherwise returns FAILURE.
\emph{Limit} only allows it to execute its child node only up to the defined number of ticks and returns the status from the child node unless the limit is reached.
\emph{Repeater} repeats its child node execution up to the input number.
\emph{Inverter} inverts the status of the child node returned, SUCCESS to FAILURE and FAILURE to SUCCESS.
\emph{Succeeder} always returns SUCCESS regardless of its child's status.
\emph{UntilFail} executes its child node until the child node returns a FAILURE.
Lastly, status manipulators such as \emph{RunningIsFailure}, \emph{RunningIsSuccess}, \emph{FailureIsSuccess}, \emph{FailureIsRunning}, \emph{SuccessIsRunning}, and \emph{SuccessIsFailure}, manipulate and propagate upward the status of its child according to the mapping defined in its node name.

\subsubsection{Composite}

Composite node is a node that contains more than one node to execute.
DeepSim Behaviour Tree supports default composite types --- \emph{selector} and \emph{sequence}.
\emph{Selector} will execute its children in order and will return SUCCESS on first SUCCESS from its child's execution.
If all of its children fail, then it returns a FAILURE statue.
\emph{Sequence} will execute its children in order and will only return SUCCESS when all its children return SUCCESS.
If any of its child nodes fails, then it returns FAILURE status.
On top of these default composite node types, DeepSim Behaviour Tree also provides parallel sequence, parallel selector, random sequence, and random selector for variation of behaviour control.

Parallel sequence executes all child nodes in parallel, and if all child nodes succeed then it stops with a SUCCESS status.
If any child node fails, then it stops with a FAILURE status.
Parallel selector also executes all child nodes in parallel, and if any child node succeeds, then it stops with a SUCCESS status.
If all child nodes fail, then it stops with a FAILURE status.
Random selector and sequence operate similarly to original selector and sequence, but the order of the child nodes' execution are shuffled for each new execution.

\subsection{DeepSimEnvs}
Most of the components in DeepSim Toolkit are focused on quickly building new environment dynamics and creating complex tasks.
Meanwhile, DeepSimEnvs component is focused on providing a quick adaptation to release newly created environments as reinforcement learning environments. 
The environment developers can implement two simple interfaces, \emph{agent} and \emph{area}, to make their environment to be released as a Unified Distributed Environment (UDE) \cite{woong2022ude} compatible environment (see Appendix \ref{appendix_deepsimenv} for the environment usage). 
The complete DeepSim component stack diagram from Gazebo to agent policy trainer is illustrated in Figure \ref{fig:stack}.

\section{Experiments}

We evaluated our two synchronization methods, \emph{get} and \emph{set}, between the environment application and Gazebo over ROS using DeepSim Toolkit Gazebo Plugin and OSRF \emph{gazebo\_ros} plugin.
As discussed in Section \ref{section:deepsim_gazebo_plugin}, OSRF \emph{gazebo\_ros} plugin can only process single object \emph{get} and \emph{set} state per service request.
Our DeepSim Gazebo Plugin overcomes this limitation by supporting \emph{get} and \emph{set} for multiple object states through a single service request.

We measured the time performance of single scene synchronization where the scene contains multiple objects.
We evaluated the performance of synchronization by incrementing ten objects up to one hundred objects to understand the impact.
For each of the synchronization calls, all objects in the scene are retrieved through \emph{get} method, and all objects positions are updated through \emph{set} method.
We measured the total time of \emph{get} or \emph{set} methods during each of scene synchronization.

\begin{figure}
    \centering
    \begin{subfigure}{0.49\textwidth}
        \centering
        \includegraphics[width=\textwidth]{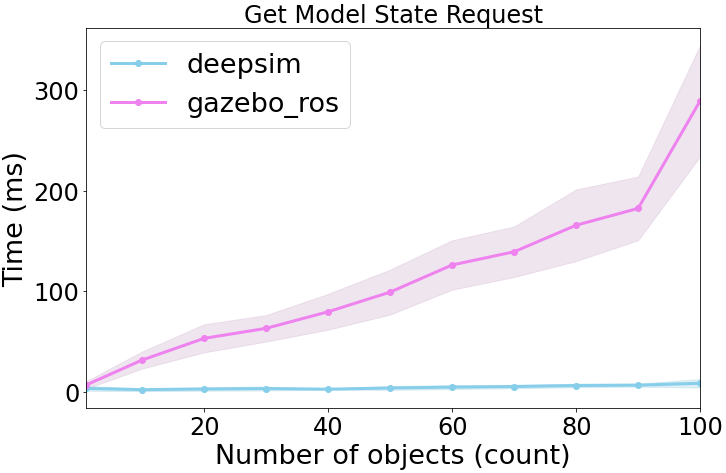}
        \label{fig:ddpg_pendulum}
    \end{subfigure}
    \begin{subfigure}{0.49\textwidth}
        \centering
        \includegraphics[width=\textwidth]{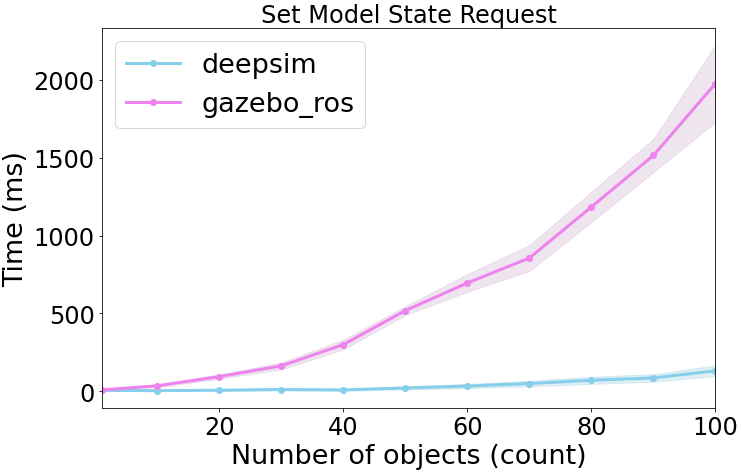}
        \label{fig:ppo_pendulum}
    \end{subfigure}
    \caption{The average time of \emph{get} and \emph{set} model state synchronization request for both DeepSim Gazebo Plugin and OSRF \emph{gazebo\_ros} Plugin}
    \label{fig:deepsim_vs_gazebo_ros_rtt}
\end{figure}

In Figure \ref{fig:deepsim_vs_gazebo_ros_rtt}, we present the average time of 5,000 synchronization requests using DeepSim Gazebo Plugin and OSRF \emph{gazebo\_ros} Plugin.
The solid curves correspond to the average time and the shaded regions correspond to the standard deviation of the request time over 5,000 synchronization requests.
DeepSim Gazebo Plugin provides a consistent average time regardless of number of objects in the scene, while \emph{gazebo\_ros} plugin's time increases significantly with respect to the number of objects in the scene.
This shows that the DeepSim Gazebo Plugin can efficiently synchronize the objects in the scene with the minimum network latency over OSRF \emph{gazebo\_ros} plugin.





\section{Conclusion}
In this paper, we described the design of the DeepSim Toolkit to build a new complex and challenging task in ROS and Gazebo with Python. 
We have designed DeepSim Toolkit to help on-board ML researchers to robotics platforms, and allow them to create more practical, complex, and challenging tasks such as autonomous driving or robotic manipulation to further expand the robotics research with ML and RL. 
We hope this toolkit brings new initiatives for both robotics and machine learning research

\bibliography{citation}
\bibliographystyle{unsrt}

\newpage
\appendix

\section{gazebo\_ros vs DeepSim Gazebo Plugin}
\label{appendix_gazebo_get_set}

\emph{gazebo\_ros} plugin provides functionality to \emph{get} and \emph{set} model state as shown in the code below. However, it only allows single retrieval or modification request of one model or link state at a time.
\begin{python}
import rospy 
from gazebo_msgs.msg import (
    ModelState,
    Pose, Quaternion, Twist
)
from gazebo_msgs.srv import SetModelState, GetModelState

# set and get ros service
set_model_state = rospy.ServiceProxy(
    "/gazebo/set_model_state",
    SetModelState)
get_model_state = rospy.ServiceProxy(
    "/gazebo/get_model_state",
    GetModelState)

# single set for agent0
state_msg = ModelState("agent0",
                       Pose(Position(1, 2, 3), Quaternion()),
                       Twist())
set_model_state(state_msg)

# another set for agent1
state_msg = ModelState("agent1",
                       Pose(Position(4, 5, 6), Quaternion()),
                       Twist())
set_model_state(state_msg)

# get agent0 and agent1 state
agent0_state = get_model_state("agent0", "")
agent1_state = get_model_state("agent1", "")
\end{python}

DeepSim Gazebo Plugin overcomes this limitation by introducing multi-state \emph{set} and \emph{get} services through a single request.
\begin{python}
from deepsim import (
    GetModelStateTracker, SetModelStateTracker,
    ModelState, 
    Pose, Position, Quaternion, Twist
)

# set multiple model states through single service request
model_states = [ModelState("agent0", 
                           Pose(Position(1, 2, 3), Quaternion()), 
                           Twist(),
                ModelState("agent1", 
                           Pose(Position(4, 5, 6), Quaternion()), 
                           Twist()]
SetModelStateTracker.get_instance().set_model_states(model_states)

# get multiple model states through single service request
model_names = ["agent0", "agent1"]
model_states = \
    GetModelStateTracker.get_instance().get_model_states(model_names)

\end{python}

\section{Environment Usage}
\label{appendix_deepsimenv}
Developers can use DeepSim Toolkit to build complex tasks by implementing \emph{AbstractAgent} and \emph{AreaInterface}. The area implementation can be passed in as an argument to create an environment. This environment provides \emph{reset} and \emph{step} interface for reinforcement learning training and evaluation.

\begin{python}
from deepsim_envs import AbstractAgent, AreaInterface

# Define robot agent behaviour.
class RobotAgent(AbstractAgent):
    def __init__(self, name):
        self._name = name
        ...
        
    def get_next_state(self):
        # return current observation of the agent.
        ...
        
    def on_action_received(self, action):
        # act upon received action command.
        ...

# Define robot area logic.
class RobotArea(AreaInterface):
    def __init__(self):
        self._agents = [RobotAgent("agent0"), RobotAgent("agent1")] 
    
    def get_agents(self):
        return self._agents
        
    def get_info(self):
        # return area info.
        ...
        
    def reset(self):
        # reset area.
        ...
    
    def observation_space(self):
        # return agents' observation spaces
        ...
    
    def action_space(self):
        # return agents' action spaces
        ...
\end{python}

\begin{python}
from deepsim_envs import Environment

# Instantiate UDE compatible environment.
robot_area = RobotArea()
env = Environment(area=robot_area)

# env provides a unified interface for reset, and step.
env.reset()
for _ in range(100):
    # sample random actions for agent0 and agent1
    action_dict = {"agent0": env.action_space["agent0"].sample(),
                   "agent1": env.action_space["agent1"].sample()}
    # submit agents' next actions and retrieve observation, reward, 
    # done, and action of all agents along with environment
    # information.
    state, reward, done, action, info = env.step(action_dict)
\end{python}

\end{document}